\documentclass{article}

\usepackage[final]{nips_2018}

\usepackage[utf8]{inputenc} % allow utf-8 input
\usepackage[T1]{fontenc}    % use 8-bit T1 fonts
\usepackage{hyperref}       % hyperlinks
\usepackage{url}            % simple URL typesetting
\usepackage{booktabs}       % professional-quality tables
\usepackage{amsfonts}       % blackboard math symbols
\usepackage{nicefrac}       % compact symbols for 1/2, etc.
\usepackage{microtype}      % microtypography

\usepackage{times}
\usepackage{latexsym}

\usepackage{times}  %Required
\usepackage{helvet}  %Required
\usepackage{courier}  %Required
\usepackage{url}  %Required
\usepackage{graphicx}  %Required
\usepackage{times}
\usepackage{latexsym}
\usepackage{algorithm}
\usepackage{algpseudocode}
\usepackage{appendix}

\usepackage{amssymb}
\usepackage{amsfonts}       % blackboard math symbols
\usepackage{microtype}      % microtypography
\usepackage{graphicx}
\usepackage{booktabs}       % professional-quality tables
\usepackage{caption}
\usepackage{subcaption}
\usepackage{amsmath}
\usepackage{color,soul}
\usepackage{multirow}
\usepackage{pbox}
\usepackage{todonotes}
\usepackage[T1]{fontenc}
\usepackage[font=small,labelfont=bf,tableposition=top]{caption}
\DeclareCaptionLabelFormat{andtable}{#1~#2  \&  \tablename~\thetable}

\newcommand{\ignore}[1]
{
}
\makeatletter
\@ifdefinable{\modelname}{\def\modelname/{\texttt{EKGN}}}
\@ifdefinable{\generalmodelname}{\def\generalmodelname/{\texttt{EKGNs}}}
\title{Link Prediction using Embedded Knowledge Graphs}

\author{Yelong Shen\footnotemark[3]\hspace{0.04in}\thanks{\hspace{0.02in}Equal contribution.}\And Po-Sen Huang\footnotemark[3]\hspace{0.04in}\footnotemark[1]\And Ming-Wei Chang\footnotemark[4]\hspace{0.04in}\thanks{\hspace{0.02in}Work performed at Microsoft Research.}\And Jianfeng Gao\footnotemark[3]\\
	\footnotemark[3]\hspace{0.09in}Microsoft Research;
	\footnotemark[4]\hspace{0.09in}Google Research\\
	\texttt{\{yeshen, pshuang, jfgao\}@microsoft.com}; \texttt{changmingwei@gmail.com}
}

\begin{document}
	
	\maketitle
	
	%\renewcommand{\thefootnote}{\fnsymbol{footnote}}
	%\footnotetext[1]{These authors contributed equally.}
	%\footnotetext[2]{The work is done at Microsoft Research.}
	\begin{abstract}
		%Why should anyone read the paper?
%TODO
Since large knowledge bases are typically incomplete, missing facts need to be inferred from observed facts in a task called knowledge base completion. The most successful approaches to this task have typically explored explicit paths through sequences of triples. These approaches have usually resorted to human-designed sampling procedures, since large knowledge graphs produce prohibitively large numbers of possible paths, most of which are uninformative. As an alternative approach, we propose performing a single, short sequence of interactive lookup operations on an embedded knowledge graph which has been trained through end-to-end backpropagation to be an optimized and compressed version of the initial knowledge base. Our proposed model, called Embedded Knowledge Graph Network (\modelname/), achieves new state-of-the-art results on popular knowledge base completion benchmarks.

	\end{abstract}
	
	%\vspace{-2mm}
	\section{Introduction}
	%% Introduce KBC tasks
Knowledge bases such as WordNet~\citep{Fellbaum98},
Freebase~\citep{BEPST08}, or Yago~\citep{SuchanekKaWe07a} contain many
real-world facts expressed as triples, e.g., (\texttt{Bill Gates},
\textsc{FounderOf}, \texttt{Microsoft}). These knowledge bases are
useful for many downstream applications such as question
answering~\citep{berant-EtAl:2013:EMNLP,yih2015semantic} and
information extraction~\citep{MBSJ09}. However, despite the
formidable size of knowledge bases, many important facts are still missing.
For example, \citet{WGMSGL14} showed that 21\% of the
100K most frequent PERSON entities have no recorded nationality in a
recent version of Freebase. 
Such missing links have given rise to the open research problem of link prediction,
or knowledge base completion (KBC) \citep{NickelTrKr11}.

% What does lead to this problem?
	% change kbc to

%% one step relation
	% direct relationship
%% multiple step relation into training
	%% multiple step relationships

%% efficiency
	%% grow expentially

	%% figure for motivation of relaxing

%What have past researchers done?
%Classify past work
%Clearly identify shortcomings and limitations
%In this paper, we develop ……..by …….which………..

%% Single step -> multiple steps
Knowledge graph embedding-based methods have been popular for tackling the KBC task.
In this framework, entities and relations (links) are mapped to continuous representations, then functions of those representations predict whether the
two entities have a missing relationship.
Equivalently, predicting whether a given edge (\texttt{h}, \textsc{r}, \texttt{t}) belongs in the graph can be formulated as predicting a candidate entity \texttt{t} given an entity \texttt{h} and a relation \textsc{r}, the input pair denoted as [\texttt{h}, \textsc{r}].
%i.e. predicting the head entity $\texttt{h}$ given a triple $(?,\textsc{r},\texttt{t})$ with relation $\textsc{r}$ and tail entity $\texttt{t}$, or predicting the tail entity $\texttt{t}$ given a triple $(\texttt{h}, \textsc{r}, \texttt{?})$ with head entity $\texttt{h}$ and relation $\textsc{r}$, where $\texttt{?}$ denotes the missing entity.

% one step
Early work \cite{BUGWY13,SocherChenManningNg2013,DBLP:journals/corr/YangYHGD14a} focused on exploring different objective functions to model {\it direct} relationships between two entities such as (\texttt{Hawaii}, \textsc{PartOf}, \texttt{USA}).
% multiple step why
Several recent approaches have demonstrated limitations of prior approaches relying upon vector-space models alone~\citep{GuuMiLi15,KristinaQu16,PTransE2015EMNLP}.
For example, when dealing with {\it multi-step} (compositional) relationships (e.g., (\texttt{Obama}, \textsc{BornIn}, \texttt{Hawaii}) $\wedge$ (\texttt{Hawaii}, \textsc{PartOf}, \texttt{USA})), direct relationship-models suffer from cascading errors when recursively applying their answer to the next input \cite{GuuMiLi15}.
% multiple step approaches
Hence, recent work \cite{gardner2014incorporating,neelakantan2015compositional,GuuMiLi15,neelakantan2015compositional,PTransE2015EMNLP,das2016chains,wang2016learning,KristinaQu16, jain2016question, DeepPath} has proposed various approaches for injecting {multi-step} paths through sequences of triples during training, further improving performance in KBC tasks.

% Challenges
Although multi-step paths through sequences of triples achieve better performance than single steps, they also introduce technical challenges.
Since the number of possible paths grows exponentially with the path length, it is prohibitive to consider all possible paths at training time for knowledge bases such as FB15k \citep{BUGWY13}.
Existing approaches need to use human-designed sampling procedures (e.g., random walks) for sampling or pruning paths of observed triples in symbolic space  (i.e., directly in the knowledge graph rather than the vector space of continuous representations.)
As most paths are not informative for inferring missing relations, these approaches can be suboptimal.

%
% \begin{figure}[!t]
%	\centering
%	\includegraphics[width=0.33\textwidth]{figures/kbc_example.pdf}
%	\qquad
%	{\small
%		\begin{tabular}[b]{l|l}\hline
%			Space & Relation Path \\ \hline
%			Symbolic & \texttt{B} $\xrightarrow[]{\text{Grandfather}^{-1}}$  \texttt{Z} $\xrightarrow[]{\text{Father}}$  \texttt{Y} $\xrightarrow[]{\text{Father}}$ \texttt{A} $\xrightarrow[]{\text{Relatives}}$ \texttt{C}   \\\hline
%			Vector & \texttt{B} $\xrightarrow[]{\text{Relatives}} \texttt{?} $ $\Rightarrow$  \texttt{A} $\xrightarrow[]{\text{Relatives}}$ \texttt{C} \\	\hline
%		\end{tabular}
%	}
%	\captionlistentry[table]{A table beside a figure}
%	\captionsetup{labelformat=andtable}
%	\vspace{-2mm}
%	\caption{{\small An illustrating example between knowledge base relation paths in symbolic space and vector space. Given a toy knowledge graph, the goal is to find a missing entity \texttt{C}, given an input [\texttt{B}, \textsc{Relative}]. In this example, since A and B are closely related in vector space (\textsc{Father} of \textsc{Father} vs. \textsc{GrandFather}), a relation path between entities \texttt{B} and \texttt{C} in vector space is quite different from a path in symbolic space.}}
%
%	% TODO	Instead of connecting triples that share exactly the same entity as in the symbolic space, \generalmodelname/ update the representations and connects other triples in the vector space instead
%
%	\label{fig:kbc_example}
%\end{figure}

 \begin{figure}[!t]
	\centering
	\includegraphics[width=0.6\textwidth]{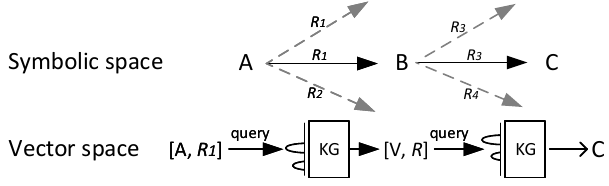}
%	\qquad
%	{\small
%		\begin{tabular}[b]{l|l}\hline
%			Space & Relation Path \\ \hline
%			Symbolic & \texttt{B} $\xrightarrow[]{\text{Grandfather}^{-1}}$  \texttt{Z} $\xrightarrow[]{\text{Father}}$  \texttt{Y} $\xrightarrow[]{\text{Father}}$ \texttt{A} $\xrightarrow[]{\text{Relatives}}$ \texttt{C}   \\\hline
%			Vector & \texttt{B} $\xrightarrow[]{\text{Relatives}} \texttt{?} $ $\Rightarrow$  \texttt{A} $\xrightarrow[]{\text{Relatives}}$ \texttt{C} \\	\hline
%		\end{tabular}
%	}
%	\captionlistentry[table]{A table beside a figure}
%	\captionsetup{labelformat=andtable}
	\vspace{-2mm}
	\caption{{\small Contrasting examples of sampling an explicit path through sequences of triples in symbolic space vs. performing a short sequence of interactive lookup operations on an embedded knowledge graph in vector space, where \texttt{A}, \texttt{B}, and \texttt{C} are entities and \textsc{r1}, \textsc{r2}, \textsc{r3}, and \textsc{r4} are relations. ``$E_{KG}$'' stands for embedded knowledge graph and [\texttt{V}, \textsc{R}] is an intermediate vector representation produced by one lookup from the entire embedded knowledge graph at once, and used to generate the query for the next lookup.}}
	\label{fig:kbc_example}
\end{figure}

In this paper, we aim to enrich neural knowledge completion models by embedding the knowledge graph through a training process.
Instead of using human-designed sampling procedures in symbolic space and training a model separately, we propose learning to perform a sequence of interactive lookup operations on the embedded knowledge graph which is jointly trained through this same process.
% Example: requirements
%Using Fig. \ref{fig:kbc_example} as a motivating example, a relation path in symbolic space ((\texttt{Z}, \textsc{Father}, \texttt{Y}) $\wedge$ (\texttt{Y}, \textsc{Father}, \texttt{A})) provides a {\it connection} between two observed triples and the {\it path length} between entity \texttt{Z} and entity \texttt{A}.
As a motivating example, consider an explicit path through sequences of triples in symbolic space ((\texttt{Obama}, \textsc{BornIn}, \texttt{Hawaii}) $\wedge$ (\texttt{Hawaii}, \textsc{PartOf}, \texttt{USA})) which provides a {\it connection} between two observed triples and a {\it path length} between the entities \texttt{Obama} and \texttt{USA}.
An equivalent traversal in vector space requires the model both to represent the connections between related triples, and to know when to stop the process.  
%Note that a ``connection'' between triples can be either sharing the same intermediate entity in a knowledge graph or semantically related.
Note that two triples can be ``connected'' by sharing either the same intermediate entity or two entities that are semantically related.
Figure \ref{fig:kbc_example} provides an illustration of the differences between traversing a knowledge graph in symbolic space vs. vector space.

%For example, in vector space, a relation path between entities \texttt{B} and \texttt{C}, [\texttt{B}, \textsc{Relatives}] $\Rightarrow$  [\texttt{A}, \textsc{Relatives}], is not constrained to share the same intermediate entity and not limited by the exponentially large number of relation paths in symbolic space if there are many relations for each entity.

%Memorize new connections
%Explore related information saved in the memory iteratively
%%Propose: embedded knowledge graph + Controller
%To address the limitations in traversing on the observed triples in symbolic space, we propose traversing on the vector space directly.
%Without directly using the connections in the symbolic space (e.g. (\textsc{A}, \textsc{Father}, \textsc{B}), (\textsc{B}, \textsc{Father}, \textsc{C})), a model needs to remember the connections between observed triples on its own to learn multi-step relations.
%In addition, since a path between triples is not given, a model needs to learn on how many steps it should traverse. I.e., a model needs to be able to adaptively change inference steps depending on the complexity of inputs.
%Though models require higher capacity to achieve the desired properties, it relaxes the constraints and resolves the limitations in traversing in symbolic space.
%An illustrating example is shown in Fig. \ref{fig:kbc_example} shows an example that traversing in vector space in more efficient than traversing in symbolic space.

\begin{figure}[t!]\centering
	%	\quad\quad\quad\quad
	{\tiny }\includegraphics[width=0.75\textwidth]{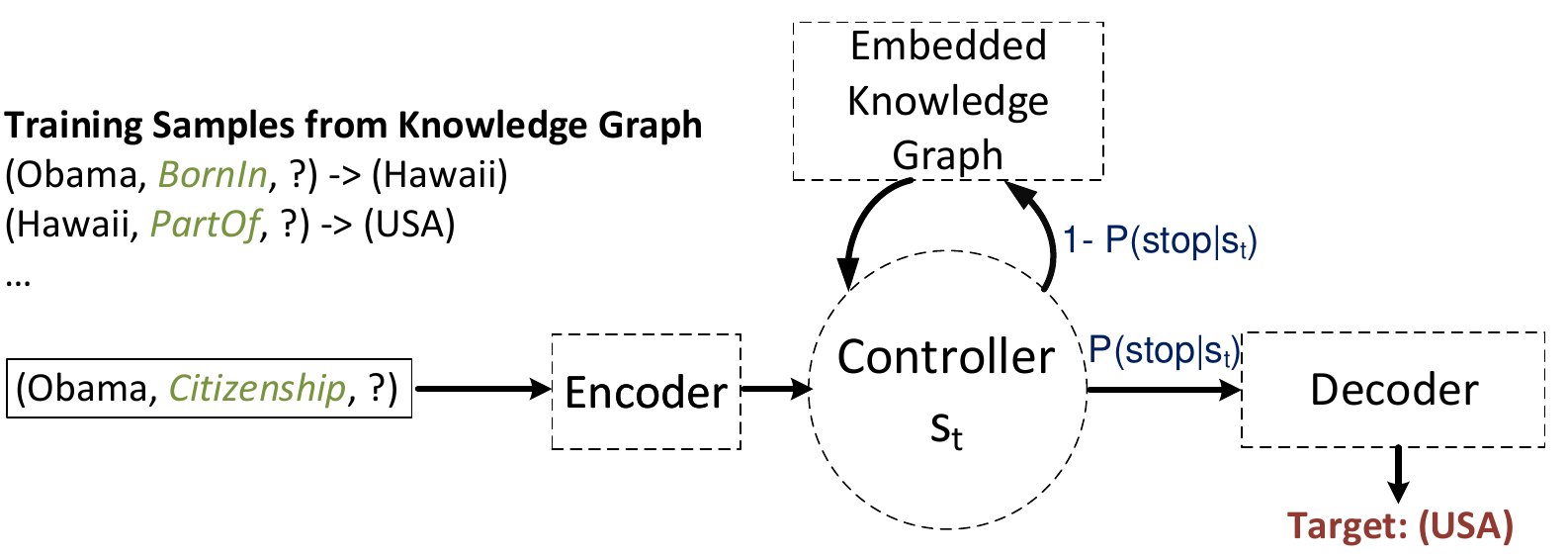} % use pdf
	\vspace{-3mm}
	\caption{	{\small An overview of the \modelname/ for KBC tasks. The embedded knowledge graph is designed to store a compact representation of the training knowledge graph, effectively learning  the connections between related triples. The controller is designed to adaptively produce lookup sequences in vector space and decide when to stop.}}
	\label{fig:architecture_overall}
\end{figure}

% how to address it
In this paper, we propose Embedded Knowledge Graph Networks (\generalmodelname/) to realize the desired properties by means of an {\em embedded knowledge graph} and a {\em controller}.
% One sentence
We design the embedded knowledge graph to learn a compact representation of the knowledge graph that captures the connections between related triples. We design the controller to learn to produce lookup sequences and to learn when to stop, as shown in Fig. \ref{fig:architecture_overall}.
% controller brief introduction
The controller, modeled by an RNN, uses a trainable termination module to decide when the \modelname/ should stop.
On each step of a lookup sequence, the controller updates its next state based on its current state and a lookup vector obtained through attention over the embedded knowledge graph.
% embedded knowledge graph brief introduction
%Without having connections in symbolic space, the design of the embedded knowledge graph is to store the connections between related observed triples in KB.

The embedded knowledge graph is an external weight matrix trained to be a compressed and optimized representation of triples in the training knowledge base.
The trained knowledge graph is ``embedded'' in the sense that it represents the original knowledge base mapped through training into a vector space.
An attention vector over the embedded knowledge graph provides context information that the controller uses to update its state and produce the next query in the lookup sequence.
% controller
%Without explicitly providing human-designed inference procedure, the design of the controller allows the model to adaptively change inference steps depending on the input and iteratively update its representation through incorporating information retrieved from the embedded knowledge graph.
%In other words, our model makes the prediction several times while forming different intermediate representations along the way.
%The controller determines how many steps the model should proceed given an input.
%At each step, a new representation is formed by taking the current representation and a context vector generated by accessing the embedded knowledge graph.
% naming
We provide interpretations and analysis of these components in the experimental section.

The main contributions of our paper are as follows: \vspace{-2mm}
\begin{itemize}
\item We propose the Embedded Knowledge Graph Network (\modelname/), which is (to the best of our knowledge) the first model that considers multi-hop relations without relying on human-designed sampling procedures or explicit paths through triples.
%to model multi-step structured relationships implicitly.
\vspace{-2mm}
\item The proposed \modelname/ can store the important information from the training knowledge graph in a relatively compact representation (e.g. 64 embedded knowledge graph vectors for 483K training triples in FB15k).
\vspace{-2mm}
\item Instead of sampling branching paths in the symbolic space of the original knowledge graph, \generalmodelname/ perform short sequences of interactive lookup operations in the vector space of an embedded knowledge graph.
\vspace{-2mm}
% update results
\item We evaluate \generalmodelname/ and demonstrate that they achieve new state-of-the-art results on the WN18, FB15k, and FB15k-237 benchmarks without using auxiliary information such as sampled paths or link features.
\vspace{-2mm}
    % respectively. Under the condition of not using structured
  % information, \modelname/ outperforms previous models by more than
  % 10\% (absolutely) in KBC tasks.
%  \todo{discover a new way of inference in models}
%TODO  update results
\item Our analysis provides insight into the inference procedures employed by \generalmodelname/.
\vspace{-2mm}
\end{itemize}

%%% Local Variables:
%%% mode: latex
%%% TeX-master: "main"
%%% End:

	%\vspace{-2mm}
	\section{Knowledge Base Completion Task}
	\label{sec:kbc}
	% KBC task
%•	Definition; Input-output of the task (Important to say it here; given that we want to claim we share the same input with other models)
    % What’s the problem?) The goal of KBC tasks (Bordes et al., 2013) is to predict a head or a tail entity given the relation type and the other entity.
%•	Formally define symbols; Talk a little bit about evaluation.
% define (h,r,t) used in the paper
The goal of Knowledge Base Completion (KBC) is to infer a missing relationship between two entities, which can be formulated as predicting a head or tail entity given the other entity and the relation type.
%i.e. predicting the head entity $\texttt{h}$ given a triple $(?,\textsc{r},\texttt{t})$ with relation $\textsc{r}$ and tail entity $\texttt{t}$, or predicting the tail entity $\texttt{t}$ given a triple $(\texttt{h}, \textsc{r}, \texttt{?})$ with head entity $\texttt{h}$ and relation $\textsc{r}$, where $\texttt{?}$ denotes the missing entity.
%Give symbol definition; how to evaluate;
%•	Say traditional way is to do inference in the symbolic space
Early work on KBC focused on learning symbolic rules. \citet{schoenmackers2010learning} learn inference rules from a sequence of triples. For instance, (\texttt{X}, \textsc{CountryOfHeadquarters}, \texttt{Y}) is implied by (\texttt{X}, \textsc{IsBasedIn}, \texttt{A}) and (\texttt{A}, \textsc{StateLocatedIn}, \texttt{B}) and (\texttt{B}, \textsc{CountryLocatedIn}, \texttt{Y}).
%o	Define symbolic space
    % Drawback in symbolic space concisely
%•	Why not using symbolic space
However, enumerating all possible relations is intractable when the knowledge base is large, since the number of distinct sequences of triples increases rapidly with the number of relation types.
Also, the rules-based methods cannot be generalized to paraphrase alternations.

% Recent work in embedding
More recently, several approaches \citep{BUGWY13,SocherChenManningNg2013,DBLP:journals/corr/YangYHGD14a} achieve better generalization by operating on embedding representations, where the vector similarity can be regarded as semantic similarity.
%TODO
During training, models learn a scoring function that optimizes the score of a target entity for a given triple.
In the evaluation, a triple with a missing entity
$(\texttt{h}, \textsc{r}, \texttt{?})$ or $(\texttt{t}, \textsc{r}^{-1}, \texttt{?})$ is mapped
into the vector space through embeddings, and the model outputs a
prediction vector for the missing entity.
The prediction is compared against all candidate entities to produce a list ranked by similarity to the prediction.
%The similarity between the prediction vector and all candidate entities, which results in a ranked list of the entities.
Mean rank and precision of the target entity in the ranked list are used as evaluation metrics.
In this paper, our proposed model uses the same task setup as in approaches of this type \citep{BUGWY13,SocherChenManningNg2013,DBLP:journals/corr/YangYHGD14a}.

%More recently, to infer a missing entity of a triple over longer sequence of triples, several approaches \citet{GuuMiLi15,KristinaQu16,PTransE2015EMNLP} propose different ways to augment path information by directly operating on the observed triples.

%To infer a missing triple, we need to capture structured relationships between multiple triples. However, the size of knowledge base is too large to consider all possible paths. Also, for some relations, it requires multiple triples to infer the right relations, e.g., (44th President of US, Citizenship, ?). It requires an even larger search space
%•	Say some approaches use auxiliary Input or inference to address the symbolic space problem!

%%% Local Variables:
%%% mode: latex
%%% TeX-master: "main"
%%% End:

	%\vspace{-2mm}
	\section{Proposed Model}
	\label{sec:reasoning_net}
	%Proposed Model (Example first or model first?)
% High-level Intuition
%The idea of our model is conceptually simple. Similar to all other knowledge base completion models, our models output a target entity as prediction given an input pair (e,r). The main difference of our model and other models is that we will try to make the prediction several times while forming different intermediate input vectors along the way. Given an input, the controller judges if we should use this input vector to produce the output prediction or not. If the controller agrees, we produce the current prediction as our final output. If not, a new input vector will be formed by taking current input and a context vector generated by accessing the global memory.   Then the new input will then be fed into the controller to judge if we should produce output or not. The whole process is performed repeatly until the controller stop the process.
%There are several advantages of using the controller. 1), the number of steps will be not fixed  separate the easy case from the difficult cases. 2) We do take simple input.  One at a time. 3)The design is to make global memory to store the KB information implicitly. 4) Because the new input is formed by the model; the "inference" is done through the neural space.!

%\paragraph{High Level Description}

%%
% The idea of our model is conceptually simple.  Similar to basic
% knowledge base completion models, our model outputs a target entity
% $\texttt{t}$ as prediction based on a single input triple $(\texttt{h}, \textsc{r}, \texttt{?})$. 

The overview of the proposed model is as follows.
The {\em encoder} module is used to
transform an input $[\texttt{h}, \textsc{r}]$ to a continuous
representation. For generating the prediction results, the {\em decoder} module takes the generated continuous representation and
outputs a predicted vector, which can be used to find the nearest entity embedding. The encoder
and decoder modules are also used to convert representations between symbolic space and
vector space for analysis.  % Note that our model does not take the path information
% explicitly.
%\todo{continous representation, hidden consistency}
Effectively, the {\em embedded knowledge graph} learns to remember connections between relevant observed triples, and the {\em controller} learns to adaptively generate sequences of queries to perform lookup operations on the graph.
The embedded knowledge graph is used to store a compact representation of the training knowledge graph as well as inferred connections between triples.
%The embedded knowledge graph is read-only by attention mechanism and is updated through back-propagation.
The controller uses a termination module to determine whether to perform another lookup step, or to stop and produce the output prediction.
If the decision is to continue, the controller learns to update its state representation using its previous state and a lookup vector obtained through attention over the embedded knowledge graph.
Otherwise, if the decision is to stop, the model produces an output prediction.
Note that the number of lookup steps is free to vary according to the complexity of each example.

We will introduce each component of the model, the detailed algorithm, training objectives, and motivating examples in the following subsections.

%The main differences between our model and previous proposed models is that we make the prediction several times while forming {\em multiple} intermediate continuous representations along the way.
%Given an intermediate representation, the {\em controller} judges if the representation encodes enough information for us to produce the output prediction or not.
%we should use the representation to produce the output prediction or not.

%If the controller agrees, we produce the current prediction as our final
%output. Otherwise, the {controller} generates a new continuous
%representation by taking current representation and a context vector
%generated by accessing the {\em embedded knowledge graph}.  Then the new
%presentation will be fed into the controller, and the whole
%process is performed repeatedly until the controller stops the
%process. Note that the number of steps varies
%according to the complexity of each example.

%for each example, as the optimal number of steps for each example could be different.

% Figure 1
\begin{figure}[t!]
%	\quad\quad\quad\quad
%\hspace{-8mm}
	\centering
	{\tiny }	\includegraphics[width=0.6\textwidth]{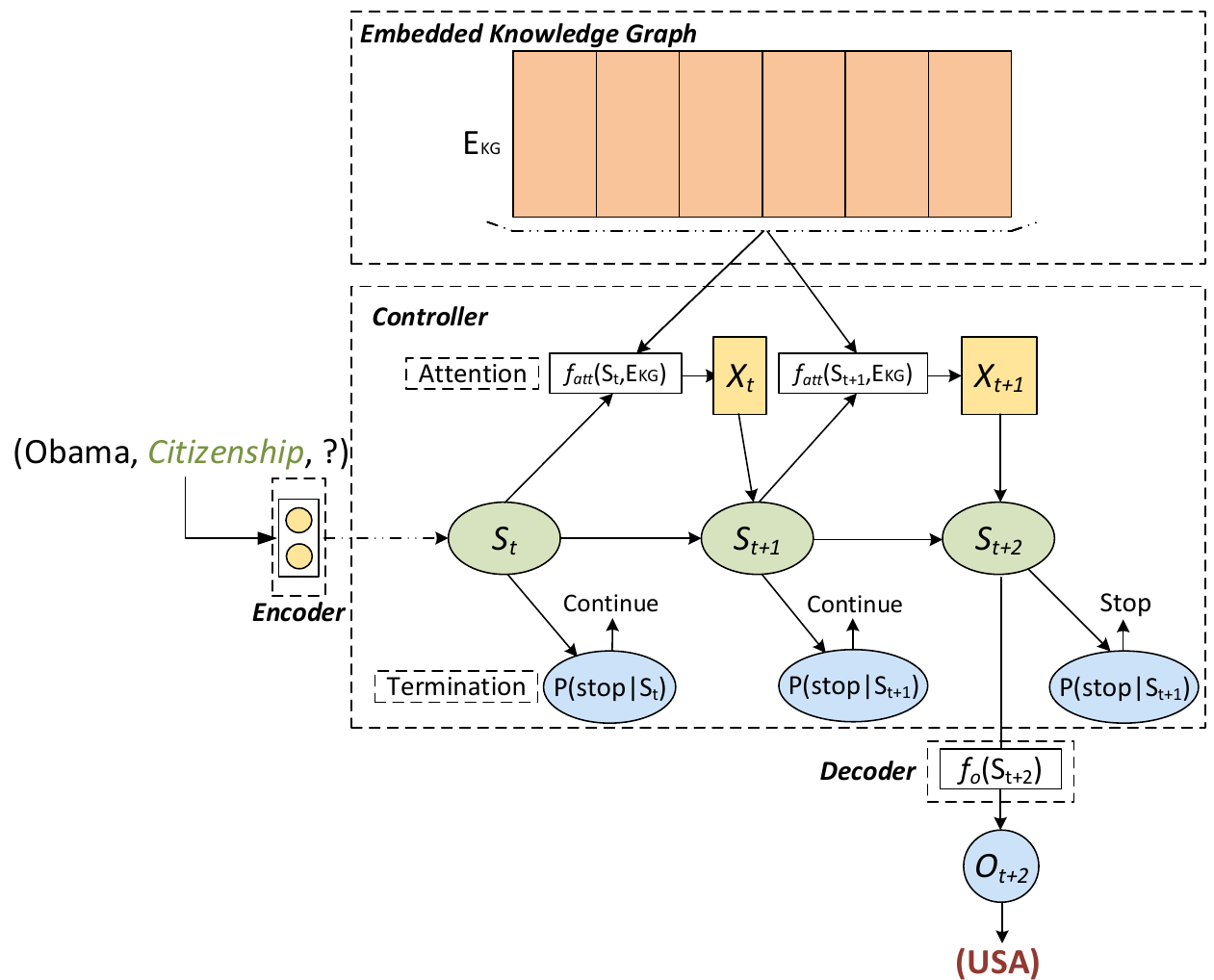} \vspace{-2mm} % use pdf
	\caption{	{\small A running example of the \modelname/ architecture. After the controller's state vector is initialized from the input $(\texttt{Obama}, \textsc{Citizenship}, \texttt{?})$, the model iteratively updates the state vector based on the current state vector and the attention vector over the embedded knowledge graph, and determines whether to stop based on the probability from the termination module.}}
	\label{fig:implicit_reasonet}
\end{figure}

\vspace{-2mm}
\paragraph{Encoder/Decoder}
%Input/Output: Input module first extracts the embedding vectors for the entity and relation from an embedding matrix.  Output: ground truth target (object) entity embedding as y
%The goal of KBC tasks \citep{BUGWY13} is to predict a head or a tail entity given the relation type and the other entity, i.e. predicting $h$ given $(?,r,t)$ or predicting $t$ given $(h, r, ?)$, where $?$ denotes the missing entity.
Given an input $[\texttt{h}, \textsc{r}]$, the encoder module retrieves the entity $\texttt{h}$ and relation $\textsc{r}$ embeddings from an embedding matrix, and then concatenates the two vectors to initialize the intermediate state representation $s_1$.

The decoder module outputs a prediction vector $f_o(s_t) = \texttt{tanh}({W_o} s_{t} + {b_o})$, a nonlinear projection from the controller hidden state $s_{t}$, where $W_o$ and $b_o$ are a weight matrix and bias vector, respectively. $W_o$ is a $k$-by-$n$ matrix, where $k$ is the dimension of the output entity embedding vector, and $n$ is the dimension of the hidden vector $s_{t}$.

\vspace{-2mm}
\paragraph{Embedded Knowledge Graph}
The embedded knowledge graph is designed to store a compact and optimized version of the information contained in the knowledge graph it is trained on, which means effectively storing the observed triples as well as semantic relations between them, in order to provide useful context information to the controller for updating its current state.
The embedded knowledge graph is denoted as $E_{KG} = \{e_i\}_{i=1}^{|E_{KG}|}$, and consists of a list of vectors that are randomly initialized.
The attention mechanism of Eq. \eqref{eq:attention} is used to access the $E_{KG}$ during both training and inference. Note that no write operations are required since the embedded knowledge graph is updated w.r.t. the training objectives in Eq. \eqref{eq:expected_reward} through back-propagation.

We now provide a motivating example for the intended functioning of the embedded knowledge graph.
Suppose, in a KBC task, that the input is [\texttt{Obama}, \textsc{Nationality}] and the model is required to output the missing entity (\texttt{USA}).
% new information
%When a new information from a new instance is received (e.g., (\texttt{Obama}, \textsc{BornIn}, \texttt{Hawaii})), the model first uses its controller to find relevant information (e.g., (\texttt{Hawaii}, \textsc{PartOf}, \texttt{U.S.A})). If the relevant information is not found, the model learns to store the information to memory vectors by gradient update in order to answer the missing entity correctly.
% Old information
%Due to the limited size of the embedded knowledge graph, the model cannot store all new information explicitly.
%Thus, the model needs to learn to utilize the embedded knowledge graph efficiently to lower the training loss.
Suppose also that the triple (\texttt{Obama}, \textsc{Nationality}, \texttt{USA}) is not contained in the knowledge graph that was used for training. The system could still produce the correct output by relying on three items of relevant information: (1) the triple (\texttt{Obama}, \textsc{BornIn}, \texttt{Hawaii}), (2) the triple (\texttt{Hawaii}, \textsc{PartOf}, \texttt{USA}), and (3) the fact that \textsc{BornIn} and \textsc{Nationality} are semantically correlated relations (i.e., nationality depends on the country where one is born).
% learn to utilize old inforamtion
Assume in this case that the controller, during training, can find items (1) and (2) in the embedded knowledge graph, but not item (3), and therefore fails to produce the right output.
The training process can then modify the embedded knowledge graph $E_{KG}$ to reflect information item (3) through back-propagation, updating the weights in such a way that the the two relations (\textsc{BornIn} and \textsc{Nationality}) are closer to each other in vector space.
In this example, similar gradient updates will have previously embedded the triple items (1) and (2) into the $E_{KG}$.
%Similarly, when the model encounters (1) or (2) for the first time, the model learns to compare  the input with the current contents of the embedded knowledge graph, and minimize the distances  between the relevant vectors in the embedded knowledge graph and the input triples. As a result, the new triple information is stored in the embedded knowledge graph indirectly by gradient update.
%Similarly, when the model encounters (1) or (2) for the first time, the model learns to compare the input with the current contents of the embedded knowledge graph, and minimize the distances between the relevant vectors in the embedded knowledge graph and the input triples. As a result, the new triple information is stored in the embedded knowledge graph indirectly by gradient update.

Limiting the size of $E_{KG}$ (to 64 in our experiments) acts as a regularizer during training, encouraging \generalmodelname/ to store information in the $E_{KG}$ in a general and reusable way.
As a result, the $E_{KG}$ becomes a compact representation of the training knowledge graph, optimized such that semantically related or similar triples, entities and relations are mapped to nearby locations in vector space, as we will illustrate using examples in Table \ref{tab:sharedMemoryVisual}.
% store new information

%iv. Search controller
\vspace{-3mm}
\paragraph{Controller}
Given an incomplete input triple, the controller is designed to coordinate a search of the embedded knowledge graph to formulate the correct output prediction, thereby lowering the training loss. To achieve this, the controller iteratively modifies its internal state representation a small but variable number of times, incorporating context information retrieved from the embedded knowledge graph at each step. The controller decides when to stop this process, at which point the output prediction is generated. Until then, the controller continues performing updates until a maximum number of steps is reached (5 in our experiments).
The controller is modeled by a recurrent neural network.

To decide when to halt this process, the controller uses a termination module to estimate
$P(\texttt{stop}|s_t)$ by logistic regression:
$\texttt{sigmoid}({W_{c}} s_t + {b_{c}})$, where the weight matrix
${W_{c}}$ and bias vector $b_{c}$ are learned during training.
With probability $P(\texttt{stop}|s_t)$ the process will be stopped
and the decoder will be used to generate the output.
With probability $1- P(\texttt{stop}|s_t)$ the controller
will generate the next representation
$s_{t+1} = \text{RNN}(s_t, x_t)$ then check the stop condition again.

The controller uses an attention
mechanism to fetch information from relevant embedded knowledge graph vectors in $E_{KG}$. The attention vector $x_t$
is generated based on the controller's current internal state
$s_t$ and the embedded knowledge graph $E_{KG}$.
% : $x_t = f_{\texttt{att}}(s_t, M)$.
Specifically, the attention score $a_{t,i}$ applied to an embedded knowledge graph vector $e_i$
given controller state $s_t$ is computed as
\begin{equation}
a_{t,i} = \texttt{softmax}_i (\lambda ~ cossim({W_1} e_i, {W_2}s_t))
\label{eq:attention}
\end{equation} where $cossim$ is the cosine similarity function, the weight
matrices ${W_1}$ and ${W_2}$ are learned during training, and $\lambda$ is chosen on the development set. ($\lambda=10$ in our experiments.) The softmax operator normalizes the attention scores across all embedded knowledge graph vectors to produce an attention distribution over $E_{KG}$ vectors. The
resulting attention vector $x_t$ is an interpolation of all $E_{KG}$ vectors, and can be written as
$x_t = f_{\texttt{att}}(s_t, E_{KG})= \sum_{i}^{|E_{KG}|} a_{t,i} e_i$.

\vspace{-2mm}
\paragraph{Overall Process}
The inference process is formally described in Algorithm \ref{alg:inference_process}.
To illustrate, given [\texttt{Obama}, \textsc{Nationality}] as input, the encoder module looks up the embedding vectors for \texttt{Obama} and \textsc{Nationality} separately, then concatenates those embeddings into a single vector $s_1$, which is the controller's initial RNN state.
Then the controller uses the probability $P(\texttt{stop}|s_t)$ to decide whether to stop and output the prediction vector $o_t$. % We can then calculate the distance between $o_i$ and embeddings of all possible entities (in the training phase, it includes the target entity and randomly selected $20$ negative entities), then convert the distance values to the probability of selecting the target entity via Eq. \eqref{eq:softmax_sample}.
Until the stop decision is made (or a predefined maximum step $T_{\text{max}}$ is reached), the controller's state $s_{t+1}$ is updated, based on its previous state $s_t$ and on the vector $x_t$ generated by performing attention over the embedded knowledge graph.
% At each step, we assign the reward of Eq. \eqref{eq:expected_reward} to be $1$ when we make a correct prediction on the target entity, and $0$ otherwise. %(i.e., $p(t|o)$ in Eq. \eqref{eq:softmax_sample})
%At test time, the model outputs
%a prediction $o_j$ where the step $j$ has the maximum termination probability.
%$j=\arg\max_k \Pi_{i=1}^{k - 1} (1-t_i) t_k$.
%\todo{Say the model is general somewhere -- done}
Note that \modelname/ is generic, and can be applied to different applications after modifying the encoder and decoder appropriately.
An example of a shortest-path synthesis task is shown in Appendix C. %\ref{sec:short-path-synth}.

%\vspace{-2mm}
\begin{algorithm}[t]
  \begin{algorithmic}
     {\small
	\State Look up entity and relation embeddings, $\mathbf{h}$ and $\mathbf{r}$.
    \State Set $s_1=[\mathbf{h}, \mathbf{r}]$ \Comment{Encode}
    \While{True}
    %\State $u \sim [0,1]$
	\If {$ u \sim P(\texttt{stop}|s_t) == 0$ and $t < T_{max}$}
	\State $x_t = f_{\texttt{att}}(s_t, E_{KG})$ \Comment {Access $E_{KG}$}
    \State$s_{t+1} = \text{RNN}(s_t, x_t)$, $t \leftarrow t + 1$ \Comment {Update RNN}
    \Else
    \State Generate output $o_t = f_o(s_t)$ \Comment{Decode}
    \State {\bf break}\Comment{Stop}
    \EndIf
    \EndWhile
	\caption{\modelname/ Inference Process}
	\label{alg:inference_process}}
  \end{algorithmic}
\end{algorithm}

%\vspace{-2mm}

%%% Local Variables:
%%% mode: latex
%%% TeX-master: "main"
%%% End:

	%\vspace{-2mm}
	\subsection{Training Objectives}
%The control is mainly on how many steps the agents should access the memory. However, the number of step is not there in the training data, so we have to use some advanced algorithm. Inspired from the REINFROCE algorithm, our job here is to optimize the expected reward using

In this section we introduce the objective function used to train \modelname/.
The number of iterations in a lookup sequence is not given
by the training data, but is learned and decided by the controller on the fly.
Therefore, the decision to terminate processing can be treated as an
action within the reinforcement learning framework.
Accordingly, the training objective is motivated by the REINFORCE algorithm \cite{Williams92}.

The expected reward at step $t$ can be obtained as follows.
% To calculate the estimated reward, we need to calculate the
% probability of the model predicting the target entity.
At step $t$, given the representation vector $s_t$, the model
generates the output vector $o_t=f_o(s_t)$.
The probability of selecting a prediction $\hat{y} \in D$
is approximated as
%\begin{equation} \label{eq:softmax_sample}
$p(\hat{y}|o_t) = \frac{\exp(-\gamma d(o_t,\hat{y}))}{\sum_{y_k \in {D}} \exp(-\gamma d(o_t, y_k)) }$,
% \end{equation}
where $d(o, y) =\| o - y
\|_1$ is the $L_1$ distance between
the output $o$ and the target entity $y$,
and $D$ is the set of all possible entities.
In our experiments, we choose hyperparameters $\gamma$ and $|D|$ based on the development set, setting $\gamma$ to 5, and sampling 20 negative examples in $D$ to speed up training.
% , respectively, for the experiments in this paper.
% In order to speed up the training, we sample a set of incorrect
% entity in $D$ instead of using all possible entities.
Defining the ground truth target entity embedding as $y^*$, the expected reward at time $t$ is defined as:
\begin{equation*}
  \begin{split}
    J(s_t|\theta) = &\sum_{\hat{y}}R(\hat{y})\frac{\exp(-\gamma d(o_t,\hat{y}))}{\sum_{\bar{y} \in {D}} \exp(-\gamma d(o, \bar{y})) }    \\
                  = & \frac{\exp(-\gamma d(o_t,y^*))}{\sum_{\bar{y} \in {D}} \exp(-\gamma d(o, \bar{y})) },
  \end{split}
\end{equation*}
where $R$ is the reward function, assigned to $1$ on a correct
prediction of the target entity, and $0$
otherwise. %(i.e., $p(t|o)$ in Eq. \eqre

Rewards are then summed over all steps. The overall
probability of the model terminating at time $t$ is $\Pi_{i=1}^{t-1} (1-v_i) v_t$,
where $v_i = P(\texttt{stop}|s_i,\theta)$. Therefore, the overall objective function can be written
as
\begin{align}\label{eq:expected_reward}
J(\theta) &= \sum_{t=1}^{T_\text{max}} \Pi_{i=1}^{t-1} (1-v_i) v_t J(s_t|\theta).
\end{align}
Given this training objective function, all parameters can be updated through back-propagation. 

	\vspace{-1mm}
	\section{Experimental Results}
	\vspace{-1mm}
	\label{sec:knowl-base-compl}
	% In this section, to demonstrate the capability of performing search over large-scale structured relationships, we first evaluate the performance of our model on knowledge-base completion tasks in Section \ref{sec:knowl-base-compl}. We further analyze the behavior of \modelname/ with a newly proposed generation task, shortest path synthesis dataset, in Section \ref{sec:short-path-synth}.
	
\begin{table}[t]
%	\centering	
	{\small
		\caption{ The knowledge base completion (link prediction) results on
			WN18, FB15k, and FB15k-237.} %\vspace{-3mm}
		\label{tab:link_prediction_wn18_fb15k}
		\hspace{-1.1cm}
		\begin{tabular}{lccccccc}
			\toprule
			Model				&	{Aux. Info.}	&	\multicolumn{2}{c}{WN18}	&	\multicolumn{2}{c}{FB15k}		&	\multicolumn{2}{c}{FB15k-237}						\\
			\cmidrule{1-8}
			&			&	Hits@10  &	MR &	Hits@10 		&	MR	&	Hits@10		&	MR	\\
%			{SE \citep{Bordes:2011:LSE:2900423.2900470}}		&	NO		&	80.5		&	 985 &	39.8	&	162	&	- & -	\\
%			{Unstructured \citep{Bordes2014}}		&	NO		&	38.2	&	304	&	6.3	&	979			&	- & - \\
			{TransE \citep{BUGWY13}}			&	NO		&	89.2	&	251	&	47.1	&	125			&	- & - \\
			{NTN \citep{SocherChenManningNg2013}}		&	NO		&	66.1	&	-	&	41.4	&	-		&	- & - \\
			{TransH \citep{Wang:2014:KGE:2893873.2894046}}		&	NO		&	86.7	&	303	&	64.4	&	87		&	- & - \\
			{TransR \citep{Lin:2015:LER:2886521.2886624}}		&	NO		&	92.0		&	225	&	68.7	&	77		&	- & -	\\
			{CTransR \citep{Lin:2015:LER:2886521.2886624}}		&	NO		&	92.3	&	218	&	70.2		&	75		&	- & -	\\
			{KG2E \citep{He:2015:LRK:2806416.2806502}}		&	NO		&	93.2	& 348	&	74.0	&	59		&	- & -	\\
			{TransD \citep{DBLP:conf/acl/JiHXL015}}		&	NO		&	92.2	&	212	&	77.3 &	91		&	- & - \\
			{TATEC \citep{DBLP:journals/corr/Garcia-DuranBUG15}}	&	NO		&	-		&	-	&	76.7	&	58 	&	- & -		\\
			{DISTMULT \citep{DBLP:journals/corr/YangYHGD14a}}	&	NO		&	94.2	&	-	&	57.7	&	-		&	41.9 & 254	\\
%			\multirow{1}{*}{STransE \citep{Nguyen2016}}		&	\multirow{1}{*}{NO}	&	94.7 (93)	&	244 ($\textbf{206}$)	&	\multirow{1}{*}{79.7} &	\multirow{1}{*}{69}		&	- & - \\
			\multirow{1}{*}{STransE \citep{Nguyen2016}}		&	\multirow{1}{*}{NO}	&	93.4	& $\textbf{206}$ &	\multirow{1}{*}{79.7} &	\multirow{1}{*}{69}		&	- & - \\
			{HOLE \citep{HOLE}}  & NO &  94.9 & - & 73.9 & -	& - & - \\
			{ComplEx \citep{trouillon2016complex}}  & NO &  94.7 & - & 84.0 & -	& - & - \\
			{TransG \citep{TransG}}  & NO &  94.9 & 345 & 88.2 & 50 	& - & - \\
			{ConvE \citep{DettmersMSR17}}  & NO &  95.5 & 504 & 87.3 & 64 	&	45.8 & 330\\
			{ProjE \citep{shi2017proje}}  & NO &  - & - & 88.4 & \textbf{34} 	&	- & - \\
			\cmidrule{1-8}
			{RTransE \citep{conf/emnlp/Garcia-DuranBU15}}		&	Path		&	-		&	-	&	76.2	&	50		&	- & -	\\
			{PTransE \citep{PTransE2015EMNLP}}		&	Path		&	-		&	-	&	84.6	&	58		&	- & -	\\
			{NLFeat \citep{Toutanova2015RepresentingTF}}		&	Node + Link Features		&	94.3	&	-	&	87.0		&	-		&	- & -\\
			{Random Walk \citep{randomwalk_kbc_emnlp16}}  & Path &  94.8 & - & 74.7 & - 	&	- & -\\
%			{NeuralLP~\citep{yang2017differentiable}}  & TensorLog &  \textbf{99.8} & - & 91.6 & - &	\textbf{49.3} & -\\
			\cmidrule{1-8}
			{\textbf{ \modelname/ }}			&		NO	&	\textbf{95.3}	& 249	& \textbf{92.7}	&	38		&	\textbf{46.4} &  \textbf{211}	\\
			\bottomrule
		\end{tabular}
	}
\end{table}

%
%\begin{table}[t]
%	\centering
%	{\small
%		\caption{The performance of \generalmodelname/ with different memory sizes and inference steps on FB15k, where $|M|$ and  $T_{max}$ represent the number of memory vectors and the maximum inference step, respectively.} \vspace{-3mm}
%		\label{tab:irn_hp_fb15k}
%		\begin{tabular}{cccc}
%			\toprule
%			%			Number of memory vectors & Maximum inference step &	\multicolumn{2}{c}{FB15k}								\\
%
%			$|M|$ & $T_{max}$ &	\multicolumn{2}{c}{FB15k}								\\
%			&	&	Hits@10 (\%) &	MR 										\\
%			\cmidrule{1-4}
%			{64} & 1		&	80.7&	55.7		\\
%			{64} & 2		&	87.4&	49.2		\\
%			{64} & {5}		&	\textbf{92.7}&	38.0		\\
%			{64} & {8}		&	88.8&	\textbf{32.9}		\\
%
%			\cmidrule{1-4}
%			{32} & {5}		&	90.1&	38.7	    \\
%			{64} & {5}		&	\textbf{92.7}&	38.0		\\
%			{128} & {5}		&	92.2&	36.1		\\
%			{512} & {5}		&	90.0&	35.3		\\
%			{4096} & {5}		&	88.7&	\textbf{34.7}		\\
%			\bottomrule
%		\end{tabular}
%	}
%\end{table}
\vspace{-3mm}

In this section, we evaluate the performance of our model on the
benchmark WN18, FB15k, and FB15k-237 datasets for KBC \citep{BUGWY13,Toutanova2015RepresentingTF}.
WN18 contains 151,442 triples with 40,943 entities and 18 relations, and FB15k consists of 592,213 triples with 14951 entities and 1,345 relations. FB15k-237 consists of 310,116 triples with 14,505 entities and 237 relations.
These
datasets contain multi-relations between head and tail entities.

Given a head (or tail) entity and a relation, a model produces a ranked list of the
entities according to the score of the entity being the tail (or head) entity of
this triple.  To evaluate the ranking, we report {\bf mean rank (MR)},
which is the mean rank of the correct entity across the test examples, and
{\bf hits@10}, which is the proportion of correct entities ranked in the top-10
predictions. Better prediction
performance is indicated by lower MR or higher hits@10. We follow the evaluation protocol in \citet{BUGWY13} in reporting filtered results, where negative examples are removed from the dataset. In this way, we prevent certain negative examples from being considered valid and ranked above the target triple.
%Following the evaluation protocol described in \citet{BUGWY13}, we remove any corrupted triples that appear in the knowledge base, to avoid cases where a correct corrupted triple \todo{define?} might be ranked above the test triple.

A single set of hyper-parameter settings is used for all datasets.
All entity and relation embedding vectors contain 100 dimensions.
The encoder module produces embeddings for input entities and relations.
The decoder module produces embeddings for output entities, which are not shared with the encoder's embeddings for those same entities.
The embedded knowledge graph consists of 64 real-valued vectors of 200 dimensions each, initialized randomly with unit $L_2$-norm.
The recurrent controller is a single-layer 200-dimensional GRU.
The maximum number of lookup steps, $T_{max}$, is set to 5.
All trainable parameters are initialized randomly.
The training algorithm is SGD with mini-batch size of 64, and learning rate 0.01. To prevent the model from learning a trivial solution by increasing entity embedding norms, we
follow \citet{BUGWY13} in constraining the $L_2$-norm of the entity
embeddings to be 1.
The validation metric for \modelname/ is hits@10.
Following \citet{PTransE2015EMNLP}, we add reverse relations into the training triple set to increase training data, i.e., for each triple $(h, r, t)$, we build two training instances, $(h, r, t)$ and $(t, r^{-1}, h)$. %# the symbols are not clearly defined

Following \citet{Nguyen2016}, we divide the results of previous work
into two groups. The first group contains the models that directly
optimize a scoring function for the triples in a knowledge base
without using auxiliary information. The second group contains the models that make use
of auxiliary information from multi-step relations. For example, the RTransE
\citep{conf/emnlp/Garcia-DuranBU15} and PTransE
\citep{PTransE2015EMNLP} models extend the TransE
\citep{BUGWY13} model by explicitly exploring multi-step relations in the
knowledge base to regularize the trained embeddings.
NLFeat \citep{Toutanova2015RepresentingTF} is a log-linear model that
makes use of simple node and link features.

%TODO update results
Table \ref{tab:link_prediction_wn18_fb15k} presents the experimental results.
According to the table, our model achieves new state-of-the-art results without using auxiliary information. Specifically, on FB15k, the MR of our model surpasses all previous
results by 12, and our hit@10 outperforms others by 5.7\%.
%On WN18, the \modelname/ obtains the highest hit@10 while maintaining similar
%MR results compared to previous work.
%\footnote{\citet{Nguyen2016}
%  reported two results on WN18, where the first one is obtained by
%  choosing to optimize hits@10 on the validation set, and second one
%  is obtained by choosing to optimize MR on the validation set. We
%  list both of them in Table \ref{tab:link_prediction_wn18_fb15k}.}
%
We also report the results of \generalmodelname/ with different embedded knowledge graph sizes $|E_{KG}|$ and different $T_{max}$ on FB15k in Appendix A. %\ref{appendix:kbc_memery_steps}.

\begin{table}[th!]
	%	\centering
	{\small
		\caption{Interpretation of state $s$ at each step, obtained by finding the closest (entity, relation) tuple, and the corresponding top-3 predictions. {``Rank''} stands for the rank of the target entity, and ``Term. Prob.'' stands for termination probability.}
%		\vspace{-3mm}
		\hspace{-1.1cm}
		\label{tab:visualization}
		\begin{tabular}{cccll}
			\toprule
			\multicolumn{5}{l}{\textbf{Input}: [\texttt{Milwaukee Bucks}, \textsc{/basketball\_roster\_position/position}]}     \\
			\multicolumn{5}{l}{\textbf{Target}: \texttt{Forward-center}}           \\
			Step       & Term. Prob.    & Rank         & \multicolumn{2}{l}{Top 3 Entity, Relation/Prediction} \\
			\cmidrule{4-5}
			\multicolumn{1}{c}{\multirow{7}{*}{1}}    & \multirow{7}{*}{6.85e-6}    & \multirow{7}{*}{5} & \multirow{3}{*}{[Entity, Relation]} & 1. [\textbf{\texttt{Milwaukee Bucks}}, \textbf{\textsc{/basketball\_roster\_position/position}}]  \\
			&                          &                     &        & 2.  [\texttt{Milwaukee Bucks}, \textsc{/sports\_team\_roster/position}] \\
			&                          &                      &       &  3.  [\texttt{Arizona Wildcats men's basketball},   \\
			&                          &                      &       &    \quad \textsc{/basketball\_roster\_position/position}]   \\
			\multicolumn{1}{c}{}        &          & & \multirow{3}{*}{Prediction} & 1. \texttt{Swingman} \\
			&                          &            &                 & 2. \texttt{Punt returner}  \\
			&                          &             &                & 3. \texttt{Return specialist}
			\\\midrule

			\multicolumn{1}{c}{\multirow{7}{*}{2}}    & \multirow{7}{*}{0.012}    & \multirow{7}{*}{4} & \multirow{3}{*}{[Entity, Relation]} & 1. [(\textbf{\texttt{Phoenix Suns}}, \textbf{\textsc{/basketball\_roster\_position/position}}]  \\
			&                          &                     &        & 2.  [\texttt{Minnesota Golden Gophers men's basketball},  \\
			&                          &                     &        & \quad \textsc{/basketball\_roster\_position/position}] \\
			&                          &                      &       &  3.  [\texttt{Sacramento Kings}, \textsc{/basketball\_roster\_position/position}]   \\
			\multicolumn{1}{c}{}       &         &  & \multirow{3}{*}{Prediction} & 1. \texttt{Swingman}    \\
			&                          &          &                   & 2. \texttt{Sports commentator } \\
			&                          &           &                  & 3. \texttt{Wide receiver} \\\midrule
			\multicolumn{1}{c}{\multirow{7}{*}{3}}    & \multirow{7}{*}{0.987}    & \multirow{7}{*}{1} & \multirow{3}{*}{[Entity, Relation]} & 1. [\textbf{\texttt{Phoenix Suns}}, \textbf{\textsc{/basketball\_roster\_position/position}}]  \\
			&                          &                     &        & 2.  [\texttt{Minnesota Golden Gophers men's basketball}, \\
			&                          &                     &        & \quad \textsc{/basketball\_roster\_position/position}] \\
			&                          &                      &       &  3.  [\texttt{Sacramento Kings}, \textsc{/basketball\_roster\_position/position}]   \\
			\multicolumn{1}{c}{}       &         &  & \multirow{3}{*}{Prediction} & 1. \textbf{\texttt{Forward-center}}    \\
			&                          &          &                   & 2. \texttt{Swingman} \\
			&                          &           &                  & 3. \texttt{Cabinet of the United States} \\
			\bottomrule
		\end{tabular}
	}
\end{table}

%\subparagraph{Illustration}

%\todo{change table 5, change analysis. Have a discuss section for why, how machine do this way. Add equations here. Potentially discuss connection in shortest path tasks.}
%In the \generalmodelname/ model, the hidden state $s_t$ is updated after accessing the global memory.
In order to explore the inference procedure learned by \generalmodelname/, we map the representation $s_t$ back to human-interpretable entity and relation names in the KB.
In Table  \ref{tab:visualization}, we show a randomly sampled example with its top-3 closest observed inputs [\texttt{h}, \textsc{r}] in terms of $L_2$-distance, and top-3 answer predictions along with the termination probability at each step.
%After accessing the global memory, the target entity has a higher rank and a higher termination probability based on the updated hidden state compared with the initial hidden state.
These mappings seem to reflect an inference procedure that is quite unlike the paths through sequences of triples in symbolic space employed by prior work  \citep{schoenmackers2010learning}.
One potential explanation is that \generalmodelname/ exercise greater flexibility by operating in the vector space of the embedded knowledge graph.
Instead of being constrained to connect only triples that share exactly the same entities in symbolic space, \generalmodelname/ can update the representations themselves and connect other, semantically related triples in vector space instead.
As shown in Table \ref{tab:visualization}, the model reformulates the representation $s_t$ at each step and gradually increases the ranking score of the correct tail entity, generating progressively higher termination probabilities during the inference process.
In the last step of Table \ref{tab:visualization}, the closest tuple (\texttt{Phoenix Suns}, \textsc{/basketball\_roster\_position/position}) is actually within the training set with a tail entity \texttt{Forward-center}, which is the same as the target entity.
These findings suggest that the \modelname/ inference process iteratively reformulates the representation $s_t$ in order to
minimize the distance between the decoder's output and the target entity in vector space.

To investigate what the model learned to store when trained on FB15k,
we randomly selected six vectors from the embedded knowledge graph,
and computed the average attention scores of each relation type for each vector.
Table \ref{tab:sharedMemoryVisual} shows the (8) top-scoring relations for each of the six vectors.
Most of a vector's top relations seem clustered around a common theme, described in the last row,
illustrating how these embedded knowledge graph vectors are activated by certain semantic patterns within the knowledge graph.
This suggests that the embedded knowledge graph can capture the semantic connections between triples.
But the patterns are broken by a few noisy relations in each column, e.g., the ``bridge-player-teammates/teammate'' relation in the ``film'' vector column, and ``olympic-medal-honor/medalist'' in the ``disease'' vector column.
%TODO
% Add appendix
We provide more \modelname/ prediction examples at each step from FB15k in Appendix B. % \ref{appendix:shared_memory}.
%In addition to the KBC tasks, we construct a synthetic task, shortest path synthesis, to evaluate the inference capability without any human-designed sampling procedure as shown in Appendix. % \ref{sec:short-path-synth}.

\begin{table}[t!]
%	\centering
	{\small
		\caption{Embedded knowledge graph visualization for an \modelname/ trained on FB15k. Each column shows the top 8 relations (ranked by their average attention scores) of one randomly selected vector in the embedded knowledge graph. 	%The first row in each column represents the interpreted relation.
		}
%		\vspace{-2mm}
		\hspace{-1.1cm}
		\label{tab:sharedMemoryVisual}
		\begin{tabular}{lll}
			\toprule
			lived-with/participant			&   person/gender							&	film-genre/films-in-this-genre		\\
			breakup/participant	      		&	person/nationality						&	film/cinematography		\\
			marriage/spouse	          		&   military-service/military-person		&	cinematographer/film		\\
			vacation-choice/vacationer		&	government-position-held/office-holder	&	award-honor/honored-for		\\
			support/supported-organization	  		&   leadership/role							&	netflix-title/netflix-genres		\\
			marriage/location-of-ceremony	&   person/ethnicity						&	director/film		\\
			canoodled/participant	        &   person/parents							&	award-honor/honored-for		\\
			dated/participant				&   person/place-of-birth 					&	bridge-player-teammates/teammate	 \\
			{\bf (Related to ``family'')} 			&   {\bf (Related to ``person'')}  					&	{\bf (Related to ``film'', ``award'')}		\\
			\cmidrule{1-3}
			disease-cause/diseases					&	sports-team-roster/team &	tv-producer-term/program \\
			crime-victim/crime-type	& basketball-roster-position/player	& tv-producer-term/producer-type \\
			notable-person-with-medical-condition/condition	&	basketball-roster-position/player	&tv-guest-role/episodes-appeared-in \\
			cause-of-death/parent-cause-of-death					&	baseball-player/position-s	&			tv-program/languages\\
			disease/notable-people-with-this-condition			&	appointment/appointed-by	&			 	tv-guest-role/actor \\
			olympic-medal-honor/medalist					&	batting-statistics/team	&					tv-program/spin-offs	\\
			disease/includes-diseases				&	basketball-player-stats/team	&	award-honor/honored-for	\\
			disease/symptoms						&	person/profession	&		tv-program/country-of-origin	\\
			{\bf (Related to ``disease'')}					& 	{\bf (Related to ``sports'')}	& {\bf (Related to ``tv program'')}	\\
			\bottomrule 
		\end{tabular}
	}
\end{table}

	\vspace{-2mm}
	\section{Related Work}
	%\vspace{-2mm}
	% What most KBC methods are doing
%     They are optimizing functions to different combinations vectors and squares.
%     at test time, find the amx minimizing the function. Given some examples,
%     
%     Talk a little about PTransE
%           Algorithms for finding path and use that te re
%
%     Structured regularization; percy; sampling
% 
%     Kristina all the on the dynamic programming; check  (need to read)
%
%
% Talk about neural turing machine
%         Have a phrase to read/and/write the structures; we do not
%         Another similarity; the planing phrase (need to read)      
%
\vspace{-2mm}
\subsection{Link Prediction using Multi-step Relations} 
\vspace{-2mm}
%Given
%that $\textsc{r}$ is a relation, $\texttt{h}$ is the head entity, and $\texttt{t}$ is the tail
%entity, most of the embedding models for link prediction focus on
%finding the scoring function $f_\textsc{r}(\texttt{h}, \texttt{t})$ that represents the
%implausibility of a
%triple by capturing direct relationship of each triple~\citep{BUGWY13,Wang:2014:KGE:2893873.2894046,DBLP:conf/acl/JiHXL015,Nguyen2016}. 
%In many studies, the scoring function $f_\textsc{r}(\texttt{h}, \texttt{t})$ is linear or bi-linear. 
%For example, in TransE~\citep{BUGWY13}, the function is
%implemented as $f_\textsc{r}(\texttt{h}, \texttt{t}) = \|\mathbf{h} + \mathbf{\textsc{r}} -\mathbf{t}\|$,
%where $\mathbf{h}$, $\mathbf{\textsc{r}}$ and $\mathbf{t}$ are the
%corresponding vector representations.
% Interestingly, studies have
% shown that complicated functions do not necessary imply good
% results~\citep{Nguyen2016, SocherChenManningNg2013}.

Recently, several studies~\cite{gardner2014incorporating,neelakantan2015compositional,GuuMiLi15,neelakantan2015compositional,PTransE2015EMNLP,das2016chains,wang2016learning,KristinaQu16, jain2016question, DeepPath} have demonstrated the importance of models incorporating multi-step paths through sequences of triples during training. Learning from multi-step
relations injects the structured relationships between triples into
the model. However, this also poses the technical challenge of
considering exponential numbers of multi-step relationships. Prior
approaches addressed this issue by designing path-mining algorithms
\citep{PTransE2015EMNLP} or considering all possible paths using a
dynamic programming algorithm with the restriction of using linear or
bi-linear models only~\citep{KristinaQu16}.
\citet{neelakantan2015compositional} and \citet{das2016chains} use an RNN to model multi-step relationships over a set of random walk paths on the observed triples.
\citet{ToutanovaCh15} shows
the effectiveness of using simple node and link features that encode structured information on FB15k and WN18.
%\citet{jain2016question} constructs memory based on relevant facts, which are extracted based on n-gram matches.
%Whereas our proposed model adaptively learns to construct the global memory and learns to query relevant information from the global memory.
\citet{DeepPath} and \citet{das2017go} use RL to traverse a knowledge graph in symbolic space, which can be viewed as an example of a human-designed sampling procedure over a knowledge graph.
%\generalmodelname/ consider the whole knowledge graph at once (instead of sampling one path in symbolic space) and traverse directly in vector space.
%Empirically, \modelname/ only traverses up to 5 steps, while the DeepPath model \citep{DeepPath} uses up to 25 steps on the FB15k-237 dataset.
%Note that we cannot compare the performance directly as the DeepPath model need to train a model for each relation, which does not scale up to knowledge graph with many relations.

% TODO rewrite
In our work, \modelname/ outperforms prior results\footnote{We cannot compare the performance directly with \citet{DeepPath} as that would require training one model for each relation, which does not scale up to a knowledge graph with a large number of relations.}
and shows that multi-step relationships can be captured by interactive lookup operations on an embedded knowledge graph.
\modelname/ adaptively learns to construct the embedded knowledge graph, and learns to look up relevant information from the embedded knowledge graph.
Each \modelname/ lookup operation potentially draws upon the entire embedded knowledge graph, instead of sampling a single path.
%Our model can be regarded as a recursive function that iteratively
%update the representation in such a way that its distance to the target entity in the neural space is minimized.
%update the representation to the target representation,
%i.e., $\|f_{\modelname/}(\mathbf{h}, \mathbf{r}) -\mathbf{t}\|$.

%\todo{Add our difference here for comparison -- done}
%Knowledge base completion is more than just link prediction.
%%TODO add
%Studies such as \citep{RYMM13, ijcai2017-166} show that incorporating textual information or context features can further improve the KBC tasks.
%It would be interesting to incorporate the information outside the
%knowledge bases in our model in the future.
%%

\vspace{-2mm}
\subsection{Neural Frameworks}
\vspace{-2mm}
%Clearly identify shortcomings and limitations
% Talk about sequence to sequence model
%Sequence-to-sequence models \citep{sutskever2014sequence,cho2014learning} have shown to be successful in many
%applications such as machine translation and conversation
%modeling~\citep{SGABJMNGD15}.  While sequence-to-sequence models are
%powerful, recent work has shown the necessity of incorporating
%an external memory to perform inference in simple algorithmic
%tasks~\citep{GravesWaDa14,GWRHDGCGRAo16}.

%TODO check others
%Discussion with respect with other framework
%A working example is shown in Figure 1.  (I would suggest to separate this figure into two; it is a bit confusing if it is the first figure. 1. We do not want to give people impression that we *directly* encode the KG. 2. We want to be careful to say that we only send one tuple at a time to the model)
\generalmodelname/ share certain features with Memory Networks (MemNN)~\citep{WestonChBo14,key_value_memnn_emnlp16,jain2016question} and Neural Turing Machines (NTM) \citep{GravesWaDa14,GWRHDGCGRAo16}. The distinctions of \generalmodelname/ are found in its controller and in its usage of the embedded knowledge graph (which corresponds to external memory in MemNN and NTM).
% problem: compare to MemNN - they have global memory as well (key-value MemNN)
MemNN and NTM explicitly store embedded inputs (such as graph triples or supporting facts) in their external memories. In contrast, \modelname/ stores no information in the embedded knowledge graph directly or explicitly. Instead, the embedded knowledge graph is initialized with random vectors,
then trained (jointly with the controller) to implicitly learn a compact and (task-specific) optimized representation
of the structured relationships in the training knowledge graph.
%We adapt the stopping mechanism in \citet{Reasonet16} to dynamically perform a multi-step inference depending on the complexity of the input.

We adapt the stopping mechanism from \citet{Reasonet16}.
Compared to \citet{Reasonet16}, \generalmodelname/ adaptively accumulate and aggregate information in the embedded knowledge graph, %Therefore, the embedded knowledge graph is forced to compress useful information across training triples (e.g. $|M|$=64 vs. 483K training triples in FB15k).
whereas \citet{Reasonet16} only constructs a paragraph embedding matrix by using the embedding of each word in an article. The paragraph embedding is not shared across samples.

	\vspace{-2mm}
	\section{Conclusion}
	\vspace{-2mm}
	%State chief result
%Higher level qualitative comments if possible 
% TODO rewrite, polish
In this paper, we propose Embedded Knowledge Graph Networks (\generalmodelname/) that learn to build and query an embedded knowledge graph.
Without using auxiliary information, the embedded knowledge graph learns to store large-scale structured relationships from the training knowledge graph in a compact way, optimized for the training task,
while the controller jointly learns to update its representations, generate lookup sequences, and decide when to stop and output the prediction.
% depending on the complexity of the input.ad
We demonstrate and analyze the \modelname/ inference process in KBC tasks.
%TODO update results
Our model, without using any auxiliary knowledge base information, achieves new state-of-the-art results on the WN18, FB15k, and FB15k-237 benchmarks.
In future work, we aim to further develop \generalmodelname/ for downstream applications such as knowledge base question answering.
	\section*{Acknowledgments}
	We thank Ricky Loynd for helpful comments and discussion, and Scott Wen-Tau Yih, Kristina Toutanova, Jian Tang, Greg Yang, Adith Swaminathan, Xiaodong He, and Zachary Lipton for their thoughtful feedback.
	%\vspace{-3mm}
	%\small
	\bibliography{cited}
	\bibliographystyle{acl_natbib}

	\clearpage
	%\null\clearpage\newpage
	\appendix

	\section{Analysis: Embedded Knowledge Graph Size and Inference Steps in KBC}
	\label{appendix:kbc_memery_steps}
	To provide additional insight into the behavior of \generalmodelname/, Table \ref{tab:irn_hp_fb15k} reports the results on FB15k for \generalmodelname/ using different hyperparameter settings. Note that an \modelname/ with $T_{max}$ = 1 is equivalent to having no embedded knowledge graph.
The table shows that the two performance metrics behave differently over hyperparameter settings.

Larger embedded knowledge graphs (to some extent) and larger maximum numbers of inference steps (to a significant extent) consistently improve the MR score, but the best hit@10 is obtained by $|E_{KG}|= $ 64 and $T_{max}$ = 5.
For both metrics, performance is much more sensitive to $T_{max}$ than it is to $|E_{KG}|$.
One possible reason is that the optimal $|E_{KG}|$ is more strongly determined by the complexity of a given task than is $T_{max}$.

\begin{table}[h]
	\centering
	{\small
		\caption{The performance of \generalmodelname/  on the FB15k test set, using different numbers of vectors in the embedded knowledge graph ($|E_{KG}|$), and different maximum numbers of inference steps ($T_{max}$). Hyperparameter settings were not selected based on these results; they were selected based on development set results.} %\vspace{-3mm}
		\label{tab:irn_hp_fb15k}
		\begin{tabular}{cccc}
			\toprule
			%			Number of memory vectors & Maximum inference step &	\multicolumn{2}{c}{FB15k}								\\
			
			$|E_{KG}|$ & $T_{max}$ &	\multicolumn{2}{c}{FB15k}								\\			 
			&	&	Hits@10 (\%) &	MR 										\\
			\cmidrule{1-4}
			{64} & 1		&	80.7&	55.7		\\
			{64} & 2		&	87.4&	49.2		\\
			{64} & {5}		&	\textbf{92.7}&	38.0		\\
			{64} & {8}		&	88.8&	\textbf{32.9}		\\
			
			\cmidrule{1-4}			    
			{32} & {5}		&	90.1&	38.7	    \\
			{64} & {5}		&	\textbf{92.7}&	38.0		\\
			{128} & {5}		&	92.2&	36.1		\\
			{512} & {5}		&	90.0&	35.3		\\
			{4096} & {5}		&	88.7&	\textbf{34.7}		\\
			\bottomrule
		\end{tabular}
	}
\end{table}
\vspace{-3mm}
	
	\section{Inference Steps in KBC}
	\label{appendix:shared_memory}
	For further analysis of the behavior of \generalmodelname/ over the sequence of lookup steps, Table \ref{tab:inference_process} gives more examples of
tail entity prediction. Interestingly, the rank of the correct tail entity improves consistently throughout the inference process.

\begin{table*}[h]
	\centering
	{\small
		\caption{Two inference examples from the FB15k dataset. Given the head entity and relation, the \modelname/ predictions at different steps are shown with their corresponding termination probabilities.}
		\label{tab:inference_process}
		%TODO(yeshen) 6 steps?, format update)
		\begin{tabular}{cccccc}
			\toprule
			\multicolumn{6}{l}{\textbf{Input}: [\texttt{Dean Koontz}, \textsc{/people/person/profession}]} \\
			\multicolumn{6}{l}{\textbf{Target}: \texttt{Film Producer} } \\
			%\cmidrule{1-4}
			Step  & Termination Prob. & Answer Rank & \multicolumn{3}{c}{Top-3 Predicted Entities} \\
			\cmidrule{4-6}
			1    & 0.018 & 9  & \texttt{Author} & \texttt{TV. Director} & \texttt{Songwriter} \\
			2    & 0.052 & 7  & \texttt{Actor} & \texttt{Singer} & \texttt{Songwriter}     \\
			3    & 0.095 & 4   & \texttt{Actor} & \texttt{Singer} & \texttt{Songwriter}     \\
			4    & 0.132 & 4  & \texttt{Actor} & \texttt{Singer} & \texttt{Songwriter} \\
			5   & 0.702 & 3   & \texttt{Actor} & \texttt{Singer} & \textbf{\texttt{Film Producer}} \\
			\cmidrule{1-6}
			% \cmidrule{1-6}
			% \multicolumn{6}{l}{\textbf{Input}: (\texttt{1964 Summer Olympics}, \textsc{/olympics/olympic\_games/sports})}\\
			% \multicolumn{6}{l}{\textbf{Target}: \texttt{Judo}} \\
			% %\cmidrule{1-4}
			% Step  & Termination Prob. & Rank & \multicolumn{3}{c}{Predict top-3 entities} \\
			% \cmidrule{4-6}
			%  1    & 0.018 & 498 & \texttt{Rugby union} & \texttt{Golf} & \texttt{Curling} \\
			%  2    & 0.079 & 4  & \texttt{Alpine skiing} & \texttt{Tennis} & \texttt{Archery} \\
			%  3    & 0.164 & 3   & \texttt{Alpine skiing} & \texttt{Tennis} & \textbf{\texttt{Judo}}    \\
			%  4    & 0.215 & 1  & \textbf{\texttt{Judo}} & \texttt{Alpine skiing} & \texttt{Archery} \\
			%  5   & 0.524 & 1   & \textbf{\texttt{Judo}} & \texttt{Alpine skiing} & \texttt{Archery} \\
			% \bottomrule

			%  \cmidrule{1-6}
			\multicolumn{6}{l}{\textbf{Input}: (\texttt{War and Peace}, \textsc{/film/film/produced\_by})}\\
			\multicolumn{6}{l}{\textbf{Target}: \texttt{Carlo Ponti}} \\
			%\cmidrule{1-4}
			Step  & Termination Prob. & Answer Rank & \multicolumn{3}{c}{Top-3 Predicted Entities} \\
			\cmidrule{4-6} 
			1    & 0.001 & 13 & \texttt{Scott Rudin} & \texttt{Stephen Woolley} & \texttt{Hal B. Wallis} \\
			2    & 5.8E-13 & 7 & \texttt{Billy Wilder} & \texttt{William Wyler} & \texttt{Elia Kazan} \\
			3    & 0.997 & 1   & \textbf{\texttt{Carlo Ponti}} & \texttt{King Vidor} & {\texttt{Hal B. Wallis}}    \\
			%  4    & 0.001 & 2  & {\texttt{King Vidor}} & \textbf{\texttt{Carlo Ponti}} & \texttt{Howard Hughes} \\
			%  5   & 6.8E-8 & 4  & {\texttt{William Wyler}} & \texttt{Billy Wilder} & \texttt{David O. Selznick} \\
			\bottomrule
		\end{tabular}
	}
\end{table*}

	\section{Analysis: Applying \generalmodelname/ to a Shortest Path Synthesis Task}
	\label{sec:short-path-synth}
	To further investigate the inference capabilities of \generalmodelname/, we constructed a synthetic {\em shortest path synthesis} task involving multi-step relations.

\begin{figure*}
	\begin{minipage}[c]{0.3\textwidth}
		\includegraphics[width=5cm]{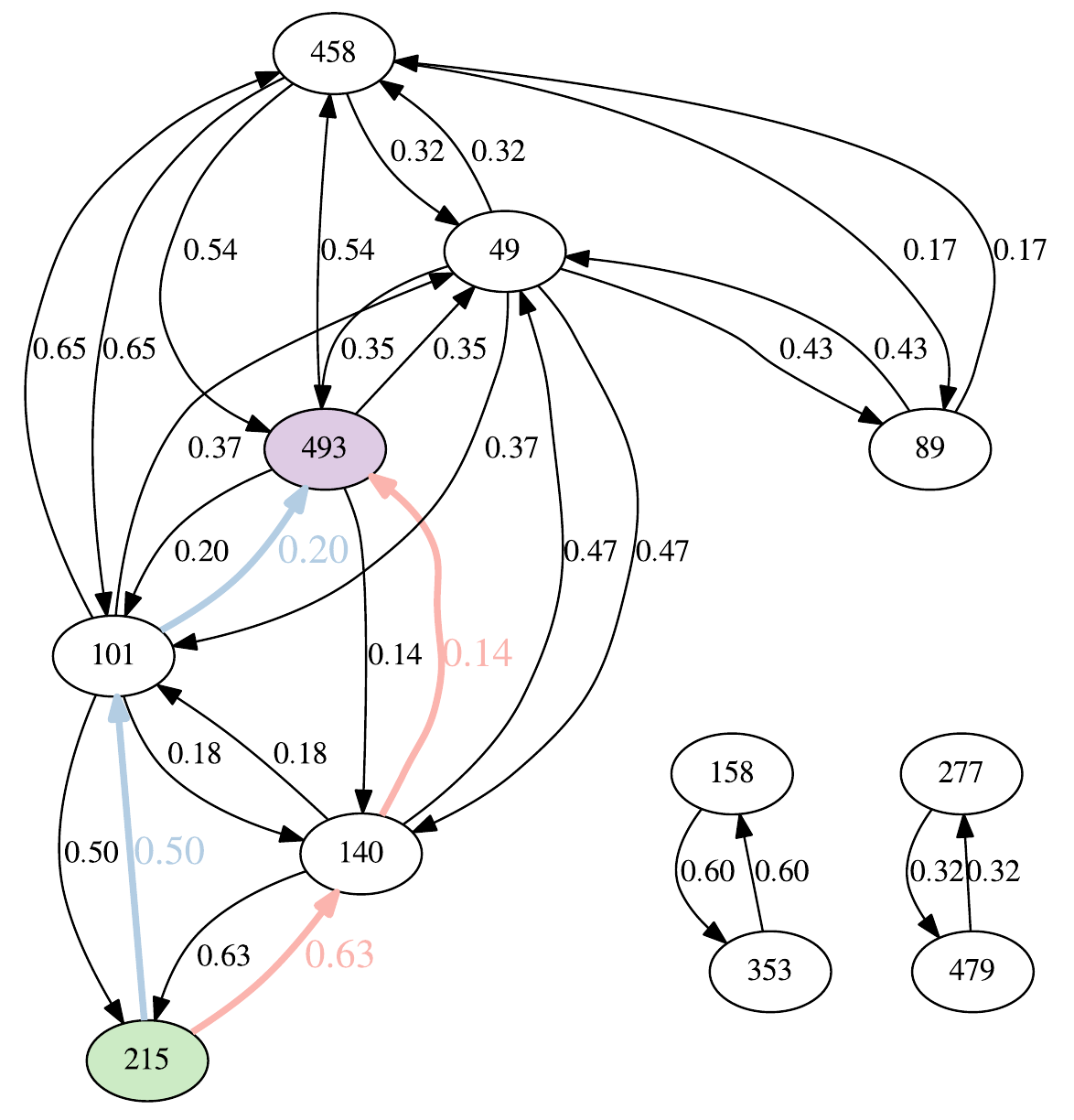}
	\end{minipage}
	%	\hfill
	\hspace{0.75cm}
	\begin{minipage}[c]{0.5\textwidth}
		{\footnotesize
			%\centering
			\begin{tabular}[b!]{c|c|c|c}\hline
				Step	&	Termination &	Distance	&	Predictions																\\
				&	Probability &				&																			\\ \hline
				1		&	0.001		&	N/A			&	$(215) \rightarrow 158 \rightarrow 89 \rightarrow 458 \rightarrow (493)$	\\
				2		&	$\sim$0			&	N/A			& 	$(215) \rightarrow 479 \rightarrow 277 \rightarrow 353 \rightarrow (493)$	\\
				3		&		$\sim$0			&	N/A			&	$(215) \rightarrow 49 \rightarrow (493)$									\\
				4		&		$\sim$0			&	0.77		&	$(215) \rightarrow 140 \rightarrow (493)$									\\
				5		&	0.999	&	0.70		&	$(215) \rightarrow 101 \rightarrow (493)$									\\\hline
			\end{tabular}
			%\captionof{table}{A table beside a figure}
		}
	\end{minipage}
	\caption{\small{An example of the shortest path synthesis dataset. Given an input ``$215 \leadsto 493$'' (Answer: $(215) \rightarrow 101 \rightarrow (493)$). Only the nodes related to this example are shown, alongside the corresponding termination probability and prediction results. The model terminates on step $5$.}}
	\label{fig:graph_visalization}
	%\end{minipage}
\end{figure*}
%\todo{
%% What is the task and how to evaluate it}
%clarification is needed:
%1. Point out explicitly that "The graph is hidden." That is, we assume that the graph is unknown.
%2. What the model can observe during training is the start-end node pairs (as input) and their shortest paths (as output).
%3. are the edge weights revealed in training data?

As illustrated in Figure \ref{fig:graph_visalization}, the input for each training instance consists of a start node and an end node (e.g., $215 \leadsto 493$) within a (hidden) weighted directed graph. The output of each instance is the set of intervening nodes in the
shortest path between the given start and end nodes, e.g., $(215)\rightarrow 101 \rightarrow (493)$.
%add clarification
The edge weights of the graph determine the lengths of paths, but are not directly revealed during training.
At test time, the model receives a start node and end node, and outputs a set of one or more intervening nodes.
An output sequence is considered correct if it connects the start node
to the end node in the underlying graph, and if the cost of the
predicted path is the same as the optimal path.

%\section{Details of the graph construction for the shortest path
%	synthesis task}
%\label{sec:deta-graph-constr}
The nodes of the underlying graph were defined by randomly sampling 500 points on a three-dimensional
unit sphere. Each node was then
connected to its $k$-nearest neighbors, using the Euclidean distances
between nodes as the edge weights. ($k=50$ by default.)
The dataset was then constructed by repeatedly sampling a random pair of nodes, and finding the shortest path connecting the first node to the second.
Paths were added to the dataset only if they contained at least one intervening node. To test multi-step inference rather than simple memorization, paths were added only if they did not contain existing paths as sub-paths, and were not contained as sub-paths within existing paths.

Note that the task is very difficult and {\em cannot} be solved perfectly by
dynamic programming algorithms since the weights on the edges are not
directly revealed. To recover some of the
shortest paths at test time, the model needs to infer
the correct path from the observed instances.
For example, assume that we observe two
instances in the training data, ``$A \leadsto D$:
$(A) \rightarrow B \rightarrow G \rightarrow (D)$'' and ``$B \leadsto E$:
$(B) \rightarrow C \rightarrow (E)$''.
In order to predict the shortest path between $A$ and $E$, the model needs to infer that ``$(A) \rightarrow B \rightarrow C \rightarrow (E)$'' is a possible path
between $A$ and $E$. If there are multiple possible paths, the model
has to decide which one is shorter using statistical
information.
% [Task Creation]
% we first graph a graph
% and then create paths
% only one graph we are being very careful about creating the graph such one no path is the superset or
% subset of the other path, so no easy answer can be found
% very difficult because we do not know the weights; some cases might not even be possible
% but it is possible for some cases, given that they are from the same graph

% IRNET Setting
The dataset was split into
20,000 instances for training, 10,000 instances for validation, and
10,000 instances for testing.
For this sequence generation task, a GRU decoder (with $128$ cells) was used as the \modelname/ output module, along with a GRU controller (with $128$ cells). Reward
$R$ was assigned to $1$ for an instance if all the predicted symbols were correct and $0$
otherwise. A $64$-dimensional embedding vector was used for input
symbols. The maximum inference step $T_{\text{max}}$ was set to
$5$.

We compare the \modelname/ with two baseline approaches: dynamic
programming without edge-weight information, and a standard
sequence-to-sequence model \citep{sutskever2014sequence} using a similar number of parameters as the \modelname/.  Without
knowing the edge weights, dynamic programming recovers only 589 correct paths at test time.
The sequence-to-sequence model recovers 904 correct paths. The \modelname/ outperforms both baselines, recovering 1,319 paths.
Furthermore, 76.9\% of the paths predicted by \modelname/ are {\em valid} paths (connecting the start and end nodes of the underlying graph). In contrast, only 69.1\% of the paths predicted by the sequence-to-sequence model are valid.

% experimental results analysis
To further understand the inference process of the \modelname/, Figure
\ref{fig:graph_visalization} shows the inference process for one test
instance.  Interestingly, to
make the correct prediction on this instance, the model has to perform
fairly complicated inference.\footnote{ In the example, to find the
  right path, the model needs to search over observed instances
  ``$215 \leadsto 448$: $(215) \rightarrow 101 \rightarrow (448)$'' and
  ``$76 \leadsto 493$:
  $(76)\rightarrow 308 \rightarrow 101 \rightarrow (493)$'', and to figure out
  that the distance of ``$140\rightarrow 493$'' is longer than
  ``$101 \rightarrow 493$''. (There are four shortest paths between
  $101\rightarrow 493$ and three shortest paths between
  $140\rightarrow 493$ in the training set.)}
We observe that the model cannot find a connected path in the first three
steps. Finally, the model finds a valid path on the fourth step, and
predicts the correct shortest path sequence on the fifth step.

%%% Local Variables:
%%% mode: latex
%%% TeX-master: "main"

\end{document}